\documentclass[10pt,onecolumn,twoside,draftclsnofoot]{IEEEtran} 
\IEEEoverridecommandlockouts     
\usepackage{amsmath,amsfonts,amssymb,amsthm,epsfig,url,array,xcolor,soul}
\usepackage{breqn}

\makeatletter
\let\NAT@parse\undefined
\makeatother

\usepackage[
pagebackref=false,
bookmarks=true,bookmarksopen=true,
bookmarksnumbered=true,
breaklinks=true,
colorlinks=true,
anchorcolor=blue,citecolor=black,
urlcolor=blue,linkcolor=blue,filecolor=blue,
menucolor=blue,
]{hyperref}
\usepackage{subcaption}
\usepackage{cite} 

\usepackage{algorithm}
\usepackage{algpseudocode}

\newtheorem{lemma}{Lemma}

\newtheorem{theorem}{Theorem}
\newtheorem{problem}{Problem}
\newtheorem{remark}{Remark}

\newcommand{\eg}{e.g.\ }

\newcommand{\cE}{\mathcal{E}}
\newcommand{\cV}{\mathcal{V}}

\newcommand{\R}{\mathbb{R}}
\newcommand{\Zp}{\mathbb{Z}^+}

\makeatletter
\@ifpackageloaded{hyperref}{%
	\@ifpackageloaded{amsmath}{%
		\newcommand{\AMShreffix}[1]{%
			\expandafter\let\csname AMShreffix#1\expandafter\endcsname%
			\csname #1\endcsname%
			\expandafter\renewcommand\csname #1\endcsname{%
				\@hyper@itemfalse\csname AMShreffix#1\endcsname}}
		\AtBeginDocument{%
			\AMShreffix{equation}
			\AMShreffix{align}
			\AMShreffix{alignat}
			\AMShreffix{flalign}
			\AMShreffix{gather}
			\AMShreffix{multline}
	}}{}}{}
\makeatother

\def\figurename{Fig.}

\DeclareMathOperator*{\minimize}{minimize\ }

\title{\LARGE \bf Resilience  in multi-robot multi-target tracking with unknown number of targets through reconfiguration}

\author{Ragesh K. Ramachandran, Nicole Fronda and Gaurav S. Sukhatme
	\thanks{This work was supported in part by the Army Research Laboratory as part of the Distributed and Collaborative Intelligent Systems and Technology (DCIST) Collaborative Research Alliance (CRA). The authors are with the Department of Computer Science, University of Southern California, Los Angeles, CA 90089, USA  {\tt\small rageshku| nfronda|gaurav@usc.edu}}
}

\begin{document}
\maketitle
\thispagestyle{empty}
\pagestyle{empty}

\begin{abstract}

We address the problem of maintaining resource availability in a networked multi-robot team performing distributed tracking of unknown number of targets in an environment of interest. Based on our model, robots are equipped with sensing and computational resources enabling them to cooperatively track a set of targets in an environment using a distributed Probability Hypothesis Density (PHD) filter. We use the trace of a robot's sensor measurement noise covariance matrix to quantify its sensing quality. While executing the tracking task, if a robot experiences sensor quality degradation, then robot team's communication network is reconfigured such that the robot with the faulty sensor may share information with other robots to improve the team's target tracking ability without enforcing a large change in the number of active communication links. A central system which monitors the team executes all the network reconfiguration computations. We consider two different PHD fusion methods in this paper and propose four different Mixed Integer Semi-Definite Programming (MISDP) formulations (two formulations for each PHD fusion method) to accomplish our objective. All four MISDP formulations are validated in simulation. 

\end{abstract}

\section{Introduction and Related work}
\label{sec: intro}

\IEEEPARstart{M}{ulti-robot} Multi-Target Tracking (MRMTT) problems have triggered considerable interest  among researchers due their immense civilian and military applications~\cite{hausman2015cooperative,olfati2011collaborative,williams2015}. This upsurge of interest in {MRMTT} problems is further expedited by the development of various decentralized/distributed tracking strategies which enable robots with limited capabilities (\eg limited field of view, memory and data processing power) to collaboratively perform the tracking task efficiently and with robustness \cite{liggins2012distributed,andreetto2018distributed,damesdistributed}. A typical MRMTT framework consists of a set of static or mobile robots (``trackers'') which are spatially distributed over a  domain of interest with each robot having at least one neighboring robot in its communication range. Each robot runs a local tracking algorithm which estimates the state of the targets in the environment using the measurements received from its field of view. Apart from performing local Multi-Target Tracking (MTT), the robots disseminate their information about the targets to their neighboring robots iteratively. Consequently, each robot refines its estimate on targets' state by appropriately fusing its local target information with the information received from its neighbors. Clearly, for any distributed strategy to function, the communication graph associated with the robots should be connected. Distributed multi-target tracking strategies eliminate the need for a centralized data processing station which was otherwise necessary for the fusion of measurement data collected by the robots to estimate the state of targets present in the environment. Moreover, distributed strategies have shown improved robustness to external noise \cite{sarkar2018asymptotic}, and are resilient to failures \cite{FB-LNS}.

Some of the important challenges to be addressed while attempting to solve multi-target tracking problems are: 1) new targets may appear in the environment and existing ones may leave the environment (number of targets are varying); 2) measurements may be generated by a non-target object (clutter measurement or false alarm); and 3) a robot may fail to detect  targets in its Field Of View (FOV) (missed detection). Although different multi-target tracking techniques have been proposed in literature which can tackle these challenges (\eg Multiple Hypothesis Tracking (MHT),  Joint Probabilistic Data-Association (JPDA)) \cite{stone2013bayesian}, Random Finite Set (RFS) based Probability Hypothesis Density (PHD) filter \cite{Mahler2003} has received the most attention. We follow a PHD filter formulation in this paper. The elegant formulation of PHD filters due to Mahler \cite{Mahler2003} provides a principled way to formulating the target tracking problem as a Bayesian filtering problem. In simple terms, the PHD of a target position state RFS is the target population density in an environment whose integral over a region yields expected number of targets in that region.  Also, a consensus-oriented decentralized strategy is employed so that the robots collaboratively estimate the  number of targets in the environment \cite{battistelli2013consensus,li2018cardinality}. It is noteworthy that, unlike the traditional target tracking algorithms (\eg MHT, JPDA), the PHD filter does not track individual target tracks. Instead, it estimates the density of the targets over time rather than the motion of individual targets. However, in recent times, using labeled RFS, researchers have extended PHD filters to enable tracking of individual targets \cite{panta2009data,Papi2015}. 
	\begin{figure}
		\centering
		\includegraphics[width=\linewidth]{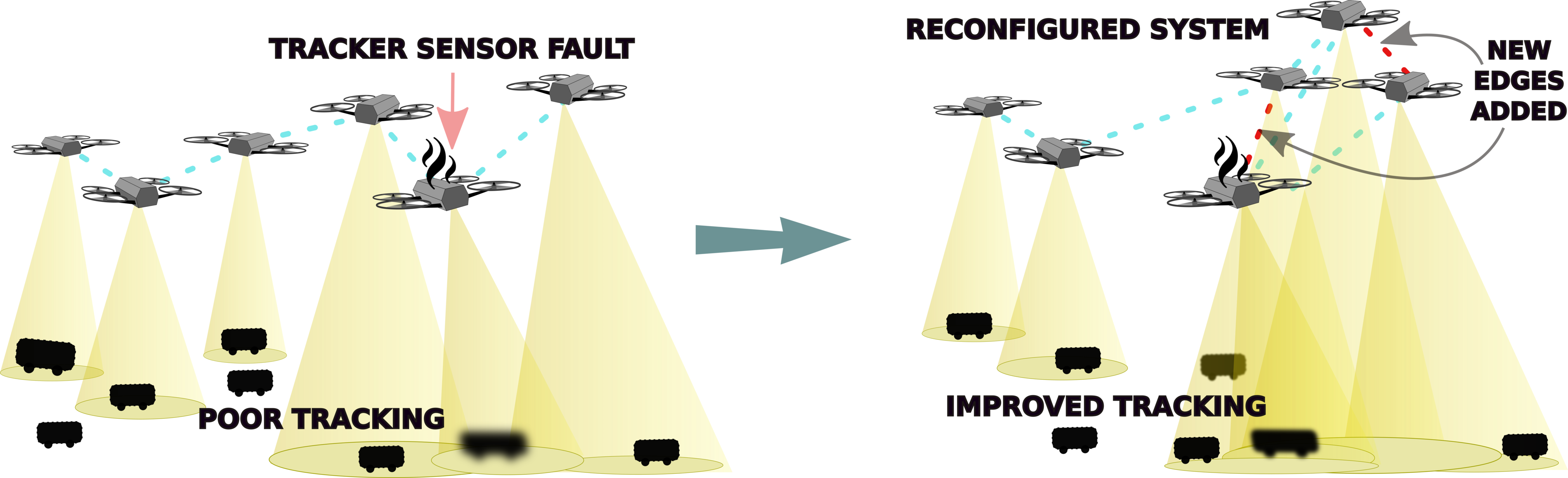}
		\caption{\small{A setting for resilient target tracking.}}
		\vspace{-9mm}
		\label{fig:failure eg}       
	\end{figure}
	
We envision a scenario in which a team of robots running a distributed PHD filter on-board cooperatively track a set of unknown number of targets steering according to some known dynamics in an environment. Moreover, the robots are overseen by a central station which intervenes with robots' tracking task only if a member in the robot team is affected by an event which result in its sensor quality deterioration. Herein, we quantify a robot's sensor quality using the trace of robot's sensor measurement noise covariance matrix. The fundamental problem we address in this paper is to attenuate the effect of a robot's sensor quality degradation on its target tracking performance (and hence the tracking performance of the robot team). We attempt to tackle this by 1) suitably reconfiguring the topology of the robot team's communication graph and 2) through optimal regeneration of   weights used for data fusion among the robots (\autoref{fig:failure eg}). We refer to the event which resulted in robot sensor quality deterioration as a \textit{detrimental event}. From a control theoretic perspective, a target tracking problem can be viewed as a state estimation problem and accuracy of state estimation is related to the observability of the system. 
The observability of networked systems with respect to its topology has gained much attention in recent times \cite{Pasqualetti2014,leitold2017controllability,Ramachandran2017}. The general consensus among researchers on this topic is that the observability of networked system in general can be improved by changing the network topology. These results motivated us to explore the possibility of improving the observability of the system through reshaping the communication graph topology, thereby mitigating the effect of sensor quality deterioration on the target tracking performance. 

Here, we extend the abstract resilience framework introduced in our previous work \cite{ramachandran2019resilience} to tackle sensor faults in the {MRMTT} setting. We had previously adopted the abstract resilience framework to handle sensor quality deterioration in the case of single target tracking \cite{ramach2019resilience}. In contrast, our current work considers the more general multi-target tracking scenario where the number of targets is time varying and unknown, the robots may receive clutter measurements or false alarms, and the targets may successfully maneuver in robots' FOV without being detected (misdetections). Following our abstract resilience framework in \cite{ramachandran2019resilience}, we mitigate the impact of a robot's sensor quality deterioration using a dual step approach. In the first step, the robot team's communication graph is modified such that the modified topology is close to the original topology and the multi-target tracking performance of the team (compared to its performance after the ``detrimental event'') is improved. We defer the details about the multi-target tracking performance metrics used in this paper to \autoref{subsec: config gen}. In addition, at this step, a set of optimal weights to fuse local PHDs among the robots in the team is also computed. The subsequent step computes a set of coordinates for the robots to embed the communication graph in the 3D space while simultaneously maximizing the robot team's coverage over the domain centered close to the centroid of the estimated targets' state PHD at that instant. Note that, our work does not consider the possibility of a complete robot sensor failure. However, under our framework, an almost-complete sensor failure can be represented as a sensor with extremely large measurement noise covariance. Moreover, we assume that the measurement noise associated with any robot's sensor is always described by a zero mean probability density function. Lastly, the robots are assumed to be able to estimate their sensor quality. This assumption does not impose any unreasonable restriction on the applicability of our strategy as several techniques exist in literature for sensor fault detection \cite{sharma2010sensor,Pierpaoli_2018}  and degradation estimation \cite{jiang2006sensor}.

Step one of our approach uses \textit{mixed integer semi-definite programs} (MISDPs) to formulate and solve the communication network reconfiguration problem and associated local PHD fusion weights generation. In this article, we consider two different kinds of local PHD fusion methods, namely: 1) \textit{Geometric Mean Fusion} (GMF) \cite{battistelli2013consensus} and 2) \textit{Arithmetic Mean Fusion} (AMF) \cite{li2018partial}. The former PHD fusion method (GMF) obtains the resultant PHD as the geometric mean of the fusing local PHDs whereas the latter computes the arithmetic mean of the fusing PHDs. In both fusing strategies, each robot also runs a consensus protocol simultaneously to distributively estimate the number of targets in the domain. The PHD fusion weights obtained from step one are employed to perform the target cardinality consensus update. Furthermore, we consider two target tracking performance criteria for each fusion method which we refer to as \textit{robot-centric} and \textit{team-centric}. As a result, we examine four MISDPs: \textit{Robot-Centric Geometric Mean Configuration generation} (\textbf{RCGMC}), \textit{Team-Centric Geometric Mean Configuration Generation} (\textbf{TCGMC}), \textit{Robot-Centric Arithmetic Mean Configuration Generation} (\textbf{RCAMC}) and \textit{Team-Centric Arithmetic Mean Configuration Generation} (\textbf{TCAMC}). In essence, a robot-centric approach optimizes the multi-target tracking performance of the robot affected by a detrimental event and team-centric approach optimizes multi-target tracking performance of the whole robot team, each with respect to a suitable metric consistent with the type of fusion rule employed.  Although resilience in multi-robot systems have received tremendous research interest \cite{Bonilla2017}, the concept of resilience through reconfiguration to improve task efficacy of the multi-robot system is recent. Through this paper, we introduce the notion of resilience by reconfiguration into multi-robot multi-target tracking.

\section{Notation and Background} 

For any positive integer $z \in \Zp$, $[z]$ denotes the set $\{1,2, \cdots, z\}$. $\|\cdot\|$ denotes the standard Euclidean 2-norm and the induced 2-norm for vectors and matrices. $\|{M}\|_F$ is the Frobenius norm of the matrix ${M} \in \R^{m_1 \times m_2}$. $Tr({M})$ is the trace of matrix ${M}$. $\bar{1}^{m}$ and $\Bar{0}^{m}$ are the vector of ones and zeros with appropriate dimensions.  We drop the superscripts whenever the dimensions of the vectors or matrices are clear from the context. $|\cdot|$ is used to denote the cardinality of a set whenever it encloses a set whereas, the same notation represents the determinant of a matrix if it encompasses a matrix. We use the same notation also to represent the number of Gaussian components in a Gaussian mixture.  Also, $\mathop{{}\mathbb{E}}[\cdot]$ represents the expectation operator. In addition, $\mathcal{N}(\bar{z}; \Bar{\mu}, \Sigma)$ denotes the Gaussian probability density of $\bar{z}$ with $\Bar{\mu}$ and $\Sigma$ representing the mean vector and covariance matrix respectively. We use ${{0}}_{\bar{i}}^{n}$ to denote a vector of zeros with one at the index $i$. 
$diag({M})$ yields the vector containing the diagonal elements of the matrix ${M}$. ${M}^{\top}$ or $({M})^{\top}$ is the transpose of ${M}$. $Blkdig(M_1, M_2, \cdots, M_n)$ outputs a block diagonal matrix with the matrices $M_1, M_2, \cdots, M_n$ along its diagonal. $\mathcal{S}^m_+$ denotes the space of $m \times m$ symmetric positive semi-definite matrices. Also, $M \succ 0$ implies $M$ is positive definite. A weighted undirected graph with non negative edge weights $\mathcal{G}$ is defined using the triplet $(\mathcal{V},\mathcal{E} \subseteq \mathcal{V} \times \mathcal{V}, \mathbf{A} \in \R^{|\mathcal{V}| \times |\mathcal{V}|}_{\geq 0})$, where ${A}$ is the weighted adjacency matrix of the graph. $\overline{\cE} = (\cV \times \cV) \setminus \cE$ is the edge complement of $\mathcal{G}$. A matrix $\mathbf{M}$ is doubly stochastic if its rows and columns sum to unity \cite{Horn:1985:MA:5509}. 

\subsection{Random finite set theory}

In this section, we review the mathematical background on random finite set theory required to understand the multi-target framework describe in this paper. A rigorous treatment on the subject can be found in \cite{mahler2007statistical, stone2013bayesian} and the references therein. A RFS is a random variable whose realizations are sets with finite cardinality.  In a multi-tracking application, RFS are used to model the set of target states and the set of measurements obtained from them at various time instants.   

A random finite set $\mathcal{X}$ can be characterized using a \textit{multi-object density function} $f(\mathcal{X})$. Unlike a random vector, the multi-object density function associated with a random set is invariant under arbitrary permutation of the elements in the set. 
Although the multi-object density function completely characterizes a RFS, using the multi-object density function for filtering application is in general intractable due the high combinatorial complexity involved \cite{Mahler2003}. Hence, approximate simpler methods are inevitable in practise. A common tractable approximation used to perform filtering on RFS is known as the \textit{probability hypothesis density}  or \textit{intensity function} filter. The probability hypothesis density function $v(\bar{x})$ is the first statistical moment of $f(\mathcal{X} = \{x_1, x_2, \cdots, x_n\})$ over the RFS $\mathcal{X}$, where the set integral \cite{mahler2007statistical,battistelli2013consensus} is applied to compute the statistical moment. An important and useful property of a PHD is that its integral over $\mathcal{R} \subseteq \mathbb{X}$ results in the expected number of targets in $\mathcal{R}$. Specifically, $\int_{\mathcal{R}} v(\bar{x}) d\bar{x} = \mathop{{}\mathbb{E}}[|\mathcal{X} \cap \mathcal{R}|]$.
To further reduce the computation burden, it is assumed that a PHD can be approximated using the following weighted finite series expansion, $v(\bar{x}) \approx \sum_{i=1}^{i_{max}} \alpha_{i} \phi_i(\bar{x})$, 
with $i_{max}$ non-negative weights $\alpha_{i}$ and basis functions $\phi_i(\bar{x})$ such that $\int_{\mathbb{X}} \phi_i(\bar{x}) d\bar{x} = 1$. It is straightforward to see that, 
\begin{align}
    \label{eqn: weight sum}
    \mathop{{}\mathbb{E}}[|\mathcal{X}|] \approx \sum_{i=1}^{i_{max}} \alpha_{i}.
\end{align}
When the Gaussian function takes the role of the basis function $\phi_i(\bar{x})$ then PHD filter is referred in literature as \textit{Gaussian Mixture } PHD (GM-PHD)\cite{panta2009data}. In the room of this paper, we use a Gaussian mixture representation for the PHD filter. This choice is driven by the fact that the GM-PHD filter equations are similar to the standard Kalman filter equations which is well suited for our MRMTT resilience framework.

\section{Problem Formulation}\label{sec:problem}

We consider a team of \textit{n} robots whose labels belong to $\{1, 2, \cdots, n\}$ tasked with tracking unknown and time-varying $\tau_k$ number of targets (at the $k^{\text{th}}$ time step). The team monitors and performs the tracking task over a compact Euclidean space $\mathcal{D}$ for a time period of $T$ epochs. However, the robots in the team can maneuver in the 3D space. To keep the computations simple, we confine $\mathcal{D}$ to be a subset of $\R^2$. Nevertheless, the formulations presented in this paper can be easily generalized to higher dimensions. Since the set of target states can be random, their collection is modeled using a {random finite set} (RFS) $\mathcal{S}_{k} = \{ \Bar{s}_1, \Bar{s}_2, \cdots, \Bar{s}_{\tau_k}\}$, where $\Bar{s}_i \in \R^4$ (position and velocity) models the state of the $i^{th}$ target that exist at time step $k$.   We refer to the robot team that tracks the moving targets as the \textit{tracker team} and the robots as  \textit{trackers}.  Let $\Bar{x}^\iota$ denotes the triplet position vector $[x^{\iota}_k, y^{\iota}_k, z^{\iota}_k] \in \R^3 $ of robot $\iota \in [n]$, then the set $\{\Bar{x}_{[n]}\}$ contains the positions of all trackers. Also, $\rho^{\iota}$ represents the robot with label $\iota \in [n]$. We assume that the trackers are equipped with localization capabilities which enable them to localize with reasonable accuracy in the environment. Since the tracker team performs Distributed Multi-Target Tracking (D-MTT) through inter-robot communication, they are equipped with resources required for communication.

 In this paper, we model the FOV of $\rho^{\iota}$ as a disc of radius $d^{\iota}_{sen}$. Now if $\rho^{\iota}$ is stationed at $\bar{x}^{\iota}_k$, then $p^{\iota}_{D,k}(\bar{s}|\bar{x}^{\iota}_k)$ denotes the probability of detection of a target with state $\bar{s}$ by $\rho^{\iota}$ at time $k$.  $p^{\iota}_{D,k}(\bar{s}|\bar{x}^{\iota}_k) = 0$ if the target (with state $\bar{s}$) lies outside the FOV of $\rho^{\iota}$ and $p^{\iota}_{D,k}(\bar{s}|\bar{x}^{\iota}_k) \leq 1$ otherwise. When $\rho^{\iota}$ flawlessly detects a target, its sensor gives a measurement $\bar{z}$ distributed according to the probability density function $h(\bar{z}|\bar{s},\bar{x}^{\iota}_k, k)$. For the computations performed in this paper, we assume that $h(\bar{z}|\bar{s},\bar{x}^{\iota}_k, k)$ can be expressed as $\mathcal{N}(\bar{z}; H^{\iota}_k\bar{s}, R^{\iota}_k)$, where ${H}^{\iota}_k$ is the sensor output matrix of $\rho^{\iota}$. ${R}^{\iota}_k$ is the covariance matrix of a Gaussian distribution modeling the sensor noise characteristics of $\rho^{\iota}$. Furthermore, we assume that $\rho^{\iota}$ receives at most one measurement per target present in its FOV at a time instant. Since the set of target measurements received by each tracker robot within its FOV is  time-varying, we use a RFS to represent the set of measurements. Let $\mathcal{Z}^{\iota}_{k} = \{\bar{z}^{\iota}_{k,1}, \bar{z}^{\iota}_{k,2}, \dots, \bar{z}^{\iota}_{k,|\mathcal{Z}^{\iota}_{k}|}\}$ denote the RFS of the set of measurements obtained by $\rho^{\iota}$ due the targets present in its FOV at time step $k$. Note that,  $|\mathcal{Z}^{\iota}_{k}|$ is less than or equal to the number of targets present in $\rho^{\iota}$'s FOV at time $k$. In addition to the measurements from the targets, a tracker may also gather false measurements due to non-target objects present in the environment. The set of false or clutter measurements acquired by $\rho^{\iota}$ is also modeled using a RFS  $\mathcal{C}^{\iota}_{k} = \{\bar{c}^{\iota}_{k,1}, \bar{c}^{\iota}_{k,2}, \dots, \bar{c}^{\iota}_{k,|\mathcal{C}^{\iota}_{k}|}\}$. Hence, the total measurements obtained by $\rho^{\iota}$ at time $k$ can be represented using the RFS $\mathcal{\Bar{Z}}^{\iota}_{k} = \mathcal{{Z}}^{\iota}_{k} \cup \mathcal{C}^{\iota}_{k}$. Let $c^{\iota}_k(\bar{z})$ denotes the PHD associated with $\mathcal{C}^{\iota}_{k}$. Furthermore, we account for new targets intruding into the domain using the RFS $\mathcal{B}_k$ and the associated PHD $b_k(\bar{s})$. Finally, $p_{S,k}(\bar{s}_{k-1})$ is the probability that a target with state $\bar{s}$ at time $k-1$ is lingering around in the environment at time $k$. In essence, $p_{S,k}(\bar{s}_{k-1})$ accounts for targets surviving in the environment. 

Let the time varying undirected graph $\mathcal{G}[k] = (\mathcal{V}, \mathcal{E}[k], {A}_u[k])$ model the communication network of the tracker team at the $k^{th}$ time step ($k \in [T]$). Note that we use ``time step'', ``time'' and ``epoch'' interchangeably in this paper. The node set $\mathcal{V}$ is isomorphic to the tracker team label set $[n]$. An edge $(i,j)$ is included in the edge set $\mathcal{E}[k]$ if $i^{th}$ robot communicates with the $j^{th}$ robot and vice versa at time $k$. We denote the communication range of trackers as $d_{mc}>0$ . The neighbor set of node $i$ in $ \mathcal{G}[k]$ is defined as $\mathcal{N}_{(i)}[k] = \{ j \in \mathcal{V}:  (i,j) \in \mathcal{E}[k] \}$. The interaction between nodes in a graph can also be represented using an unweighted adjacency matrix. The unweighted adjacency matrix of $\mathcal{G}[k]$ is denoted by ${A}_u[k]$. 

\subsection{Distributed Multi-Target Tracking (D-MTT)}
\label{subsec: distrib Kalman filter}

In general, a MTT problem can decomposed into two estimation problems: 1) estimation of the number of targets present in the environment and 2) estimation of targets' state. The PHD filter is a computationally efficient way to simultaneously solve both  estimation problems in a tractable way. If multiple trackers are employed to monitor a region of interest for intruders, then MTT can be performed distributively without the use of a centralized data fusion center. 
As mentioned earlier, in distributed multi-target tracking, each tracker runs a local PHD filter using the its measurements obtained for its field of view and transmits relevant information about its local PHD filter to its neighboring trackers. The neighboring trackers then update their local PHDs by fusing the received information with their PHDs. In this paper, we examine two different PHD fusion methods proposed in literature for our resilience framework, namely, \textit{geometric mean fusion} (GMF) \cite{battistelli2013consensus} and \textit{arithmetic mean fusion} (AMF) \cite{li2018partial}. Interestingly, both fusion strategies can be elegantly derived as optimal solutions to two different minimization problems involving the weighted Kullback-Leibler divergence (KLD) between the fusing multi-object densities \cite{abbas2009kullback}. Furthermore, AMF and GMF do not double-count information as long as their fusing weights sum to unity \cite{bailey2012conservative}. In the forthcoming subsections, we describe the local PHD filter employed by each tracker and the two different PHD fusing strategies used in this paper. 

\subsection{Tracker local PHD filter}
\label{subsec: local PHD}

Analogous to a Kalman filter, a PHD filter also consists of a \textit{prediction} step and an \textit{update} or \textit{innovation} step. In the prediction step of the PHD filter, PHD associated with the target states RFS is updated based on the target dynamics and the target birth RFS $\mathcal{B}_k$. Subsequently, in the innovation step, the targets' state RFS PHD is refined using the measurements received from the targets. We assume that every target in the environment follows the following standard linear state space dynamics equation
\begin{align}
    \label{eqn:tracking agent dynamics}
    \bar{s}_{k+1} = {F}_k\bar{s}_{k} + {G}_k\bar{u}_k + \bar{w}_k,
\end{align}
where $\mathbf{x}_{k} \in \R^{s_a}$ and  $\mathbf{u}_k \in \R^{u_a}$ are the state and the input vectors of a target respectively. ${F}_k \in \R^{s_a \times s_a}$ and ${G}_k \in \R^{s_a \times u_a}$ are the state transition matrix and input matrix of appropriate dimensions respectively. $\Bar{w}_k \in \R^{s_a}$ is the zero mean normally distributed random vector with the covariance matrix ${Q}_k \in \R^{s_a \times s_a}$ ($\Bar{w}_k \sim \mathcal{N}(\Bar{0}, {Q}_k)$).  

In D-MTT, each tracker maintains and updates a local PHD filter. The prediction and update PHD filter equation associated with a tracker $\rho^{\iota}$ can be mathematically expressed as \cite{Mahler2003,stone2013bayesian},
\vspace{-0.15in}
\begin{align}
    \label{eqn: PHD prediction}
    &v^{\iota}_{k|k-1}(\Bar{s}) =  b_k(\Bar{s}) + \nonumber \\ \int & p_{S,k}(\Bar{s}_{k-1}) f(\Bar{s}|\Bar{s}_{k-1}, \Bar{u}_{k-1})v^i_{k-1|k-1}(\Bar{s}_{k-1}) d\Bar{s}_{k-1} \\
    \label{eqn: PHD innovation}
    &v^{\iota}_{k|k}(\Bar{s}) = (1-p^{\iota}_{D,k}(\bar{s}|\bar{x}^{\iota}_k))v^{\iota}_{k|k-1}(\Bar{s}) + \nonumber \\ 
     \sum_{\Bar{\zeta} \in \mathcal{\Bar{Z}}^{\iota}_{k}} &\frac{p^{\iota}_{D,k}(\bar{s}|\bar{x}^{\iota}_k)h(\bar{\zeta}|\bar{s},\bar{x}^{\iota}_k, k) v^{\iota}_{k|k-1}(\Bar{s})}{c^{\iota}_k(\bar{\zeta}) + \int p^{\iota}_{D,k}(\bar{s}|\bar{x}^{\iota}_k)h(\bar{\zeta}|\bar{s},\bar{x}^{\iota}_k, k) v^{\iota}_{k|k-1}(\Bar{s}) d\Bar{s} },
\end{align}
 $f(\Bar{s}|\Bar{s}_{k-1}, \Bar{u}_{k-1})$ is the probability of occurrence of target state $\Bar{s}$ at time $k$ given the previous states and inputs, derived using \autoref{eqn:tracking agent dynamics}. \autoref{eqn: PHD prediction} and \autoref{eqn: PHD innovation} are the {prediction} and { update or innovation} equations respectively. 
 

Despite the fact that, in general, it is hard to further simplify the above equations, a closed form expression can be derived if the PHDs are assumed to be Gaussian Mixtures (GM), and target motion and measurement models are assumed to linear \cite{vo2006gaussian}. As indicated earlier, we adopt a GM approximation for the PHDs used in the paper. Thus $v^{\iota}_{k|k}(\Bar{s}) \approx \sum_{i=1}^{|v^{\iota}_{k|k}|} \alpha^{\iota,(i)}_{k|k} \mathcal{N}(\Bar{s}; \Bar{\mu}^{\iota,(i)}_{k|k}, P^{\iota,(i)}_{k|k})$ and $b_k(\Bar{s}) \approx \sum_{i=1}^{|b_k|} \alpha^{b,(i)}_{k} \mathcal{N}(\Bar{s}; \Bar{\mu}^{b,(i)}_{k}, P^{b,(i)}_{k})$.
Also, we prescribe that $c^{\iota}_k(\bar{\zeta}) = \lambda^{\iota}_{c,k} |FOV|^{\iota} p_{FOV}(\bar{\zeta}) $, where $p_{FOV}(\bar{\zeta})$ is the probability density function of the occurrence of clutter measurements in $\rho^{\iota}$'s FOV (assumed to be uniform in this paper), $|FOV|^{\iota}$ is the ``volume'' of its FOV and $\lambda^{\iota}_{c,k}$ is the expected number of clutter measurements per unit FOV volume. For clarity of presentation, hereafter, we restrict our attention to tracker state independent  $p^{\iota}_{D,k}(\bar{s}|\bar{x}^{\iota}_k, k)$ and $p_{S,k}(\bar{s}_{k-1})$, the formula for more general case can be found in \cite[Section III-E]{vo2006gaussian}. Under these assumptions and approximations \autoref{eqn: PHD prediction} and \autoref{eqn: PHD innovation} can be written as, 
 
 \textit{GM-PHD filter prediction:}
 \begin{align}
     \label{eqn: GM-PHD prediction}
     v^{\iota}_{k|k-1}(\Bar{s}) &= b_k(\Bar{s}) + v^{\iota}_{S,k|k-1}(\Bar{s}) \\
     \label{eqn: predicted GM-PHD}
     v^{\iota}_{S,k|k-1}(\Bar{s}) = & p_{S,k} \sum_{i=1}^{|v^{\iota}_{S,k|k-1}|} \alpha^{\iota,(i)}_{k|k-1} \mathcal{N}(\Bar{s}; \Bar{\mu}^{\iota,(i)}_{k|k-1}, P^{\iota,(i)}_{k|k-1}) \\
     \label{eqn: predicted GM mean}
     \Bar{\mu}^{\iota,(i)}_{k|k-1} &= {F}_{k-1}\Bar{\mu}^{\iota,(i)}_{k|k-1} + {G}_{k-1}\bar{u}_{k-1}\\
     \label{eqn: predicted GM covariance}
     P^{\iota,(i)}_{k|k-1} &= Q_{k-1} + F_{k-1}P^{\iota,(i)}_{k-1|k-1}F^{\top}_{k-1},
 \end{align}
\textit{GM-PHD filter innovation:}
\begin{align}
\label{eqn: GM-PHD innovation}
v^{\iota}_{k|k}(\Bar{s}) &= (1-p^{\iota}_{D,k})v^{\iota}_{k|k}(\Bar{s}) + \sum_{\Bar{\zeta} \in \mathcal{\Bar{Z}}^{\iota}_{k}} v^{\iota}_{D, k}(\Bar{s}; \Bar{\zeta}) \\
\label{eqn: GM-PHD innov detect}
v^{\iota}_{D, k}(\Bar{s}; \Bar{\zeta}) &= \sum_{i=1}^{|v^{\iota}_{D, k}|} \alpha^{\iota,(i)}_{k|k}(\Bar{\zeta}) \mathcal{N}(\Bar{s}; \Bar{\mu}^{\iota,(i)}_{k|k}(\Bar{\zeta}), P^{\iota,(i)}_{k|k}) \\
\label{eqn: GM-PHD innov weight}
\alpha^{\iota,(i)}_{k|k}(\Bar{\zeta}) = & \frac{p^{\iota}_{D,k}\alpha^{\iota,(i)}_{k|k-1}\mathcal{N}(\Bar{\zeta}; H^{\iota}_{k}\Bar{\mu}^{\iota,(i)}_{k|k-1}, S^{\iota,i}_{k})}{c^{\iota}(\bar{\zeta})+ p^{\iota}_{D,k} \sum_{j} \alpha^{\iota,(j)}_{k|k-1}\mathcal{N}(\Bar{\zeta}; H^{\iota}_{k}\Bar{\mu}^{\iota,(j)}_{k|k-1}, S^{\iota,j}_{k})} \\
\label{eqn: innovation measure covariance}
S^{\iota,i}_{k} &= R^{\iota}_{k} +  H^{\iota}_{k}P^{\iota,(i)}_{k|k-1}(H^{\iota}_{k})^{\top} \\
\label{eqn: innov mean updt}
\Bar{\mu}^{\iota,(i)}_{k|k} &= \Bar{\mu}^{\iota,(i)}_{k|k-1} + K^{\iota,(i)}_{k}(\Bar{\zeta}-H^{\iota}_{k}\Bar{\mu}^{\iota,(i)}_{k|k-1}) \\
\label{eqn: innv covariance}
P^{\iota,(i)}_{k|k} &= [I - K^{\iota,(i)}_{k}H^{\iota}_{k}]P^{\iota,(i)}_{k|k-1}\\
\label{eqn: Kalman gain}
K^{\iota,(i)}_{k} &= P^{\iota,(i)}_{k|k-1}(H^{\iota}_{k})^{\top}(S^{\iota,i}_{k})^{-1}.
\end{align}
\cite[Table 1]{vo2006gaussian} gives a pseudocode for the Gaussian mixture PHD filter implementation. Notice that \autoref{eqn: predicted GM mean}-\autoref{eqn: predicted GM covariance} and \autoref{eqn: innovation measure covariance}-\autoref{eqn: Kalman gain} are similar to the prediction and innovation steps of a standard Kalman filter respectively \cite{stone2013bayesian}. Finally, the number of Gaussian components (GCs) in $v^{\iota}_{k|k}(\Bar{s}) $ are reduced for computational efficiency by merging closer GCs and pruning GCs with low weights (see \cite[Section III.C, Table II]{vo2006gaussian}).

\subsection{Local PHD fusion}
\label{subsec: distribued PHD fusion}

In a PHD fusion method, the estimated local targets' state PHD $v^{\iota}_{k|k}(\Bar{s})$ of $\rho^{\iota}$ is fused with the estimated local targets' state PHDs of other neighboring trackers in the tracker team. From hereon, unless otherwise specified, PHD or local PHD mean local targets' state PHD. Certainly, fusing all the PHD Gaussian components (GCs) of $\rho^{\iota}$ with all the PHD Gaussian components received from its neighbours is inefficient in terms of both computational and communication load. To this end, we follow the method proposed in \cite{li2018partial}. According the strategy delineated in \cite{li2018partial}, each tracker disseminates only the highly weighted GCs or Target likely GCs (T-GCs) (that possibly corresponds to real targets) in its PHD to its neighboring trackers. In our work, we identify the T-GCs in a PHD using the \textit{rank rule} outlined in \cite{li2018partial}. Consequently, the numbers of T-GCs selected equals the integer closest to the expected number of targets according a tracker's local PHD. Recall that, from \autoref{eqn: weight sum}, the sum of the weights of a GM-PHD gives the expected number of targets in the region of interest. Let ${\alpha}^{\iota}_{k}=\sum_{i}\alpha^{\iota,(i)}_{k|k}$ and $\Tilde{\alpha}^{\iota}_{k}=\left \lceil{{\alpha}^{\iota}_{k}}\right \rceil $. A neighboring tracker can then fuse its local PHD T-GCs with the communicated T-GCs using any established fusion method of choice \cite{battistelli2013consensus,Li2019}. As noted earlier, in this article, we focus on AM and GM based fusion strategies described in latter subsections. In both cases, an additional cardinality consensus scheme is simultaneously executed and each tracker's PHD GCs are rescaled so that the tracker's estimate of the expected number targets converges to the global average. The cardinality consensus rule can be mathematically written as:
\begin{align}
    \label{eqn: consnss crdinlty updte}
    \Tilde{\alpha}^{i}_{k}(l) &= \sum_{j \in \mathcal{N}_{(i)}[k] \cup \iota} [{\Bar{A}}[k]]_{i,j} \Tilde{\alpha}^j_{k}(l-1),
\end{align}
where $[{\Bar{A}}[k]]_{i,j}$ is the $(i,j)$ entry of a doubly stochastic matrix ${\Bar{A}}[k]$ with the same structure as the unweighted adjacency matrix (${A}_u[k]$) of connected graph $\mathcal{G}[k]$ except for the diagonal elements. Specifically, $\mathbf{\Bar{A}}[k]$ is non-zero along its diagonal and its off-diagonal elements are non-zero if and only if the corresponding elements of $\mathbf{A}_u[k]$ are unity. In theory, cardinality consensus scheme shown in \autoref{eqn: consnss crdinlty updte} converges to a common quantity only when $l$ tends to infinity. However, it is known that consensus protocols enjoy an exponential rate of convergence \cite{FB-LNS}. Thus, a reasonable level of consensus on the global average can be attained by iterating \autoref{eqn: consnss crdinlty updte} for a sufficient number of consensus steps $L$. We note that, the consensus update is assumed to happen at a much faster time scale compared to the target dynamics (\autoref{eqn:tracking agent dynamics}). After each consensus step $l$, along with fusing the local T-GCs with the neighbor's T-GCs (using GMF or AMF), the PHD GCs weights of each tracker are rescaled such that they sum to  $\Tilde{\alpha}^{\iota}_{k}(l)$, i.e., $\alpha^{\iota,(i)}_{k|k}(l) = w^{\iota}\alpha^{\iota,(i)}_{k|k}(l)$, with $w{\iota} = \frac{\Tilde{\alpha}^{\iota}_{k}(l)}{\sum_{i} \alpha^{\iota,(i)}_{k|k}(l)}$.  An analysis on the accuracy of cardinality estimation in the cardinality consensus scheme can be found in \cite{li2018cardinality}. In following two subsections, we will delineate the two PHD (primarily T-GCs of the PHDs) fusion strategy adopted in our paper. 


\subsubsection{Geometric mean fusion}
\label{subsubsec: gm fusion}

Let ${d_1, d_2, \dots, d_g}$ be a set of probability density function defined over some state space, then these probability density functions can be fused into a single probability density function $d_{GCI}$ based on a set of weights $\{\omega_i \geq 0\}$ using the generalized covariance intersection fusion rule (GCI) using $d_{GCI} = C^{-1} \prod_{i \in [g]} d^{\omega_i}_i,$
%
the normalization constant $C$ is given by $\int \prod_{i \in [g]}d^{\omega_i}_i$. When $\sum_{i \in [g]} \omega_i = 1$ this fusion rule is also referred as \textit{exponential mixture density}. 

Examining formula to compute $d_{GCI}$, we can observe that if the weights sum to unity then $d_{GCI}$ is proportional to the geometric mean of the densities. Hence, we refer to the GCI fusion rule as \textit{Geometric Mean Fusion} (GMF). A detailed discussion on GCI fusion can be found \cite{battistelli2013consensus,abbas2009kullback} and the references therein. Since the PHDs can be interpreted as unnormalized probability density functions, the fused PHD can be computed similar to $d_{GCI}$ with $C=1$. 

Suppose $\sum_{i \in [g_1]} \alpha^{1}_{i} \mathcal{N}(\Bar{s}; \Bar{\mu}^{1}_{i}, P^{1}_{i}) $, $\sum_{i \in [g_2]} \alpha^{2}_{i} \mathcal{N}(\Bar{s}; \Bar{\mu}^{2}_{i}, P^{2}_{i}) $, $\cdots$, $\sum_{i \in [g_1]} \alpha^{n}_{i} \mathcal{N}(\Bar{s}; \Bar{\mu}^{1}_{i}, P^{1}_{i}) $ are the GM-PHDs to be fused together using GMF based on the normalized set of weights $\{\omega_i \geq 0 \}_1^n$.  We refer to a set of weights as \textit{normalized} if the weights sum to unity. Also, let $\varpi$ denote the dimension of the mean vectors in the GCs. Under the assumption 
that \cite{julier2006empirical}, 
\begin{align}
    \label{eqn: GM pow approx}
    \left[\sum_{i \in [g]} \alpha_{i} \mathcal{N}(\Bar{s}; \Bar{\mu}_{i}, P_{i}) \right]^{\omega} \approx \sum_{i \in [g]}\left[\alpha_{i} \mathcal{N}(\Bar{s}; \Bar{\mu}_{i}, P_{i}) \right]^{\omega},
\end{align}

we derive the following formula to compute the resultant GM-PHD ($v_{GMF}$) obtained by fusing the GM-PHDs according to GMF: 
\begin{align}
    \label{eqn: GM-PHD GMF}
    v_{GMF}=\sum_{i_1 \in [g_1]} \cdots \sum_{i_n \in [g_n]} \alpha^{1, \cdots,n}_{i_1, \cdots, i_n} \mathcal{N}(\Bar{s}; \Bar{\mu}^{1, \cdots,n}_{i_1, \cdots, i_n}, P^{1, \cdots,n}_{i_1, \cdots, i_n}) 
\end{align}
where 
\begin{align}
    \label{eqn: GMF covariance rule}
    \left(P^{1, \cdots,n}_{i_1, \cdots, i_n} \right)^{-1} &= \sum_{j=1}^n \omega_j (P^{j}_{i_j})^{-1} \\
    \label{eqn: GMF state rule}
    \left(P^{1, \cdots,n}_{i_1, \cdots, i_n} \right)^{-1} \Bar{\mu}^{1, \cdots,n}_{i_1, \cdots, i_n} &= \sum_{j=1}^n \omega_j (P^{j}_{i_j})^{-1} \Bar{\mu}^{j}_{i_j} \\
    \label{eqn: GMF coeff rule}
    \alpha^{1, \cdots,n}_{i_1, \cdots, i_n} = K&\left(\prod_{j=1}^n (\alpha_{i_j}^{j})^{\omega_j} \sqrt{\frac{\left|2 \pi \frac{{P}_{i_j}^{j}}{\omega_j}  \right|}{|2 \pi {P}_{i_j}^{j} |^{w_j}}}\right) 
\end{align}
 ${\small K = \exp{(\Tilde{K} - \Bar{K})}}$; ${\small \Tilde{K} =-\frac{1}{2}\left( n \varpi \ln{(2 \pi)}  \right.}$ $- \sum_{j=1}^n \ln{\left|\omega_j({P}_{i_j}^{j})^{-1}\right| }$  $\left. \sum_{j=1}^n \omega_j (\Bar{\mu}_{i_j}^{j})^T ({P}_{i_j}^{j})^{-1} \Bar{\mu}_{i_j}^{j} \right)$ and \newline $\Bar{K} = -\frac{1}{2}\left( \varpi \ln{(2 \pi)}  - \log |\sum_{j=1}^n w_j ({P}_{i_j}^{j})^{-1}| + \Bar{q}^T \left(\Omega\right)^{-1} \Bar{q} \right)$; with  
%
$\Omega = \sum_{j=1}^n \omega_j (P^{j}_{i_j})^{-1}$ and $\bar{q} = \sum_{j=1}^n \omega_j (P^{j}_{i_j})^{-1} \Bar{\mu}^{j}_{i_j}$. The derivation of \autoref{eqn: GM-PHD GMF} is a straightforward application of the formula to compute the product of multivariate Gaussian functions \cite{bromiley2003products} and therefore is ignored in this article. From \autoref{eqn: GM-PHD GMF}, one can conclude that the number  GCs in the fused GM-PHD exponentially increases at each fusion step, which would easily saturate storage and computational capabilities of a tracker. The common approaches devised to control the growth of GCs in GM-PHD are merging GCs which are close is some sense (usually in terms of Mahalanobis distance \cite{battistelli2013consensus,vo2006gaussian}) and pruning GCs whose weights fall below a pre-defined threshold. Here, we follow the latter and prune GCs in the fused GM-PHD which have low weights ($\alpha^{1, \cdots,n}_{i_1, \cdots, i_n} \ll 1$).

\begin{remark}
As noted earlier, an important weakness of GMF is the exponential growth of GCs in the fused GM-PHD. In addition, a more serious defect of GMF as pointed out in \cite{Yi2017} is its susceptibility to misdetections. Hence, a misdetection by single tracker could potentially jeopardize the MTT performance of other trackers significantly. 
\end{remark}

\subsubsection{Arithmetic mean fusion}
\label{subsubsec: AMF}

If one utilizes the arithmetic mean fusion (AMF) rule to fuse the probability density functions $\{d_1, d_2, \cdots, d_g\}$ defined over some space to yield a single probability density function over the state space, according to a set of normalized weights, then the resultant fused probability density function $d_{AMF}$ is given by $d_{AMF} = \sum_{i \in [g]} \omega_i d_i$.


Akin to the GMF case, the AMF rule to used to fuse probability density functions is applied to fuse GM-PHDs. Fusing $n$ GM-PHDs according to AMF yields, 
\begin{align}
    \label{eqn: AMF GM-PHD}
    v_{AMF} = \sum_{j\in[n]}\omega_j\left(\sum_{i \in [g_j]} \alpha^{j}_{i} \mathcal{N}(\Bar{s}; \Bar{\mu}^{j}_{i}, P^{j}_{i})\right).
\end{align}

According to \autoref{eqn: AMF GM-PHD}, fusing two GM-PHDs with $g_1$ and $g_2$ GCs create a GM-PHD with $g_1+g_2$ GCs whereas, the GMF of the GM-PHDs result in a GM-PHD containing $g_1 \times g_2$ GCs. Hence, in general, the number of GCs resulting from AMF is much smaller compared to the GMF case. The number of GCs in the AMF fused GM-PHD can be further reduced by fusing GCs which potentially describes the same target using the following Gaussian Mixture Reduction (GMR) technique. 

Without loss of generality, we assume that the first GC from each GM-PHD to be fused represented as $\alpha^1_1\mathcal{N}(\bar{s}; \Bar{\mu}^1_1, P^1_1)$, $\alpha^2_1\mathcal{N}(\bar{s}; \Bar{\mu}^2_1, P^2_1)$, $\cdots$, $\alpha^n_1\mathcal{N}(\bar{s}; \Bar{\mu}^n_1, P^n_1)$ describe the same target (due to their closeness to each other in some sense).  These GCs are combined and reduced to a single GC $\alpha^{GMR}_1\mathcal{N}(\bar{s}; \Bar{\mu}^{GMR}_1, P^{GMR}_1)$ using the associated normalized weights $\{\omega_1, \omega_2, \cdots, \omega_n\}$. Then \cite{li2018partial}, 
\begin{align}
    \label{eqn: AMF weight rule}
    \alpha^{GMR}_1 &= \sum_{i \in [n]} \alpha^i_1 \\
    \label{eqn: AMF state rule}
    \Bar{\mu}^{GMR}_1 &= \frac{\sum_{i \in [n]}\alpha^i_1 \Bar{\mu}^i_1}{\sum_{j \in [n]}\alpha^j_1}\\
    \label{eqn: AMF covariance rule}
    P^{GMR}_1 &= \sum_{i \in [n]} \omega^i P^i_1.
\end{align}

It has been shown in \cite{li2018partial} that, as long as each GC involved in the fusion process is ``conservative/consistent'' with respect to an estimate on the target state, the fused the GC is also conservative (see \cite{uhlmann1995dynamic} for the details on conservative/consistent estimate pair ). In other words, AMF GC avoids underestimating the actual estimate errors in mean square sense, and are resilient to misdetections \cite{uhlmann2003covariance}. 

\subsection{Tracking under Sensor Quality Deterioration}

As stated in \autoref{sec: intro}, we consider the problem of attenuating the effect of a tracker's sensor quality deterioration on multi-target tracking performance through appropriate reconfiguration of the tracker team. In this subsection, we will define the notion of tracker team reconfiguration and sensor quality deterioration used in this paper. 

We define the tuple $(\mathcal{G}[k], {\Bar{A}}[k])$ as the \textit{configuration} of the tracker team at the $k^{th}$ time step and denote it by $\mathcal{C}[k]$. ${\Bar{A}}[k]$ is a doubly stochastic matrix whose elements are used as normalized weights for the local PHD fusion, and to execute the cardinality consensus operations outlined in \autoref{eqn: consnss crdinlty updte}. During the multi-target tracking task, executed for a time period $T$, let $n_f$ detrimental events occur independently to arbitrary trackers in the team. We assume that each event results in some sensor quality deterioration. At time $k$, we say that $\rho^{\iota}$'s sensor quality is deteriorated if the trace of the measurement noise covariance matrix associated with its sensor ${R}^{\iota}_k$ has increased compared to $Tr({R}^{\iota}_{k-1})$. In more formal terms, if $Tr({R}^{\iota}_k)$ $>$ $Tr({R}^{\iota}_{k-1})$, then $\rho^{\iota}$'s sensor quality deteriorated at time $k$. Recall our assumption that, the detrimental event never introduces any bias in the tracker's sensor measurements. The treatment of detrimental event which results in sensor bias is reserved for future work. Similar to \cite{ramach2019resilience}, we consider a sequence set $\mathcal{F} = [f_1, f_2, \cdots, f_p, \cdots, f_{n_f}]$, where $f_p$ indicate the time step when the $p^{th}$ sensor fault occurred. We specify that $\mathcal{C}[f_p-1]$ is the configuration of the tracker team before the $p^{th}$ detrimental event occurred. We now formally define the problems studied in this paper. The first problem (\autoref{prob: configuration generation}) deals with reconfiguration of the tracker team such that target tracking performance is optimal in some reasonable sense. The second problem addresses the issue of realizing the graph topology in 3D space while maximizing the tracker team's coverage over the $\mathcal{D}$. 

\begin{problem}
		\label{prob: configuration generation}
		\textbf{Configuration generation or reconfiguration:}
		Given that $\rho^i$ experienced sensor quality deterioration at some time $f_p$, ${R}^i_{f^+_p}$ is the sensor noise covariance matrix immediately after the deterioration event, and $\mathcal{C}[f_p-1]$ is the tracker configuration prior to the event, determine a new configuration $\mathcal{C}[f_p]$ such that,
		\begin{enumerate}
		    \item $\mathcal{G}[f_p]$ is a connected graph, 
		    \item $\|{A}_u[f_p] - {A}_u[f_p-1] \|^2_F \leq 2\times e$, where $e \in \Zp$ is the number of edges that may be modified in $\mathcal{G}[f_p-1]$ to obtain $\mathcal{G}[f_p]$, and
		    \item tracking performance is optimized is some appropriate sense.
		\end{enumerate}
\end{problem}

\begin{problem}
        \label{prob: formation synthesis}
        \textbf{Formation synthesis:} Given a tracker team configuration $\mathcal{C}[f_p]$, generate coordinates that best realize the given configuration and maximize the tracker team's coverage over $\mathcal{D}$, subject to constraints. We defer the  details of this problem to \autoref{subsec: formation syn}.
\end{problem}

Graph connectivity constraint is  essential for the  distributed tracking computation performed over the network and thus is included in \autoref{prob: configuration generation} \cite{FB-LNS}. The second condition enables the user to control the communication load on the generated configuration by tuning the parameter $e$. The final condition assures improved multi-target tracking performance. 

\begin{figure}
		\centering
		\includegraphics[width=0.7\linewidth]{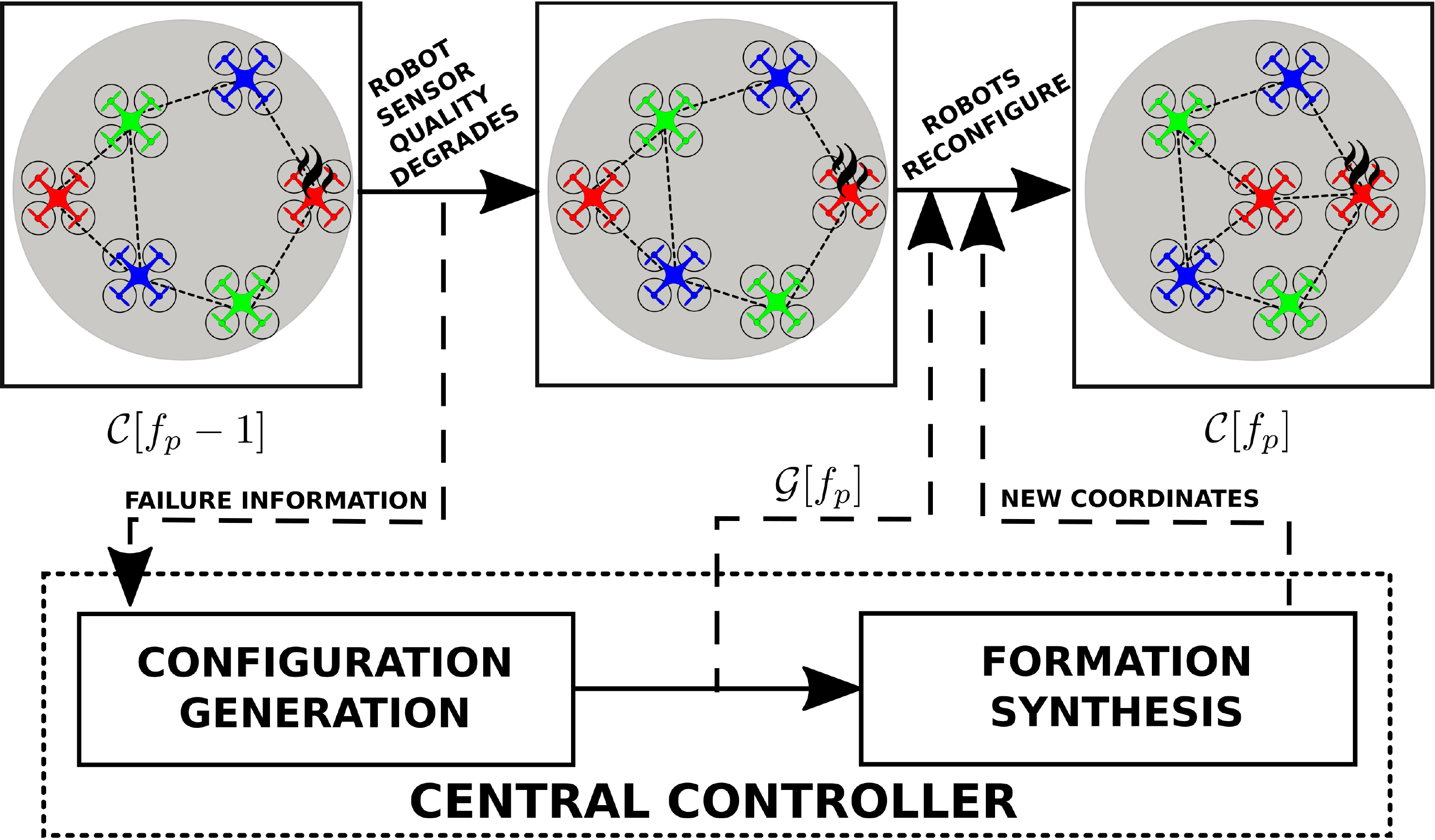}
		\caption{\small{Basic outline of our approach. When a robot experiences sensor quality degradation, \emph{configuration generation} selects edges to modify the communication graph. Then, \emph{formation synthesis} assigns robots to physical locations that support the desired graph topology. }}
		\label{fig:schematic}       
\end{figure}

\section{Methodology}
\label{sec:Methodology}

In this section, we focus on developing various strategies for solving \autoref{prob: configuration generation} and \autoref{prob: formation synthesis}. In our framework, we consider a base station that monitors the activities of the tracker team. The base station intervenes with the team's operation and instructs the multi-target tracking team only when a detrimental event occurs. Similar to \cite{ramachandran2019resilience,ramach2019resilience} , the base station controller adopts a dual-step scheme to arrive at a new tracker team configuration and compute the new tracker coordinates which best realize the new configuration in space. Akin to \cite{ramachandran2019resilience}, these steps are referred as \textit{configuration generation} and \textit{formation synthesis}. The whole decision making scheme is depicted in \autoref{fig:schematic}. In essence, the solutions to \autoref{prob: configuration generation} and \autoref{prob: formation synthesis} are the bases for the \textit{configuration generation} and \textit{formation synthesis} steps respectively.

\subsection{Configuration Generation}
\label{subsec: config gen}

In general, we classify the configuration generation strategies delineated in this paper into two types: \textit{robot-centric} and  \textit{team-centric}. A robot-centric approach aims at improving the tracking performance of the robot which endured the adverse effects of a detrimental event, whereas a team-centric approach optimizes the tracking performance of the whole team. As described earlier, in this article, we restrict our attention to two kinds of local PHD fusion strategies. Hence, for each fusion strategy, we can devise either a tracker-centric or a team-centric approach to configuration generation. We formally formulate all the four configuration generation approaches in this subsection. We refer the four configuration generation approaches as \textit{Robot-Centric Geometric Mean Configuration generation} (\textbf{RCGMC}), \textit{Team-Centric Geometric Mean Configuration generation} (\textbf{TCGMC}), \textit{Robot-Centric Arithmetic Mean Configuration generation} (\textbf{RCAMC}) and \textit{Team-Centric Arithmetic Mean Configuration generation} (\textbf{TCAMC}). All  four configuration generations result in solving different mixed integer semi-definite programs (MISDPs). The motivation for this stems from the fact that network design problems are often formulated in literature as MISDPs \cite{rafiee2010optimal}. Note that, in the MISDPs formulations we optimize the tracking performance for one step PHD fusion ($l=1$). We drop the dependence of variables on time in the MISDPs for brevity. The following theorem guides our design of the connectivity constraint in the MISDPs (See \hyperref[app: graph connect proof]{Appendix~\ref{app: graph connect proof}} for the proof). 
\begin{theorem}
\label{theorem:graph connectivity}
If a graph containing self loops at every node is equipped with a weighted adjacency matrix ${A}$ which is doubly stochastic then any graph isomorphic to this graph with or without self loops is connected if and only if $\frac{1}{n} \Bar{1} \bar{1}^{\top}  + {I}  \succ {A}$.
\end{theorem}
At time $f_p$, $\rho^{\iota}$ experienced a sensor fault and $\alpha = \min \{\Tilde{\alpha}^{1}_{f_p},\dots,\Tilde{\alpha}^{\iota}_{f_p}, \dots, \Tilde{\alpha}^{n}_{f_p}\}$. As a result of \autoref{eqn: consnss crdinlty updte}, each tracker should have at least one GC associated with a target up to $\alpha$ targets. In addition, let $\{P^{\iota}_i\}_{i=1}^{\alpha}$ be the covariance matrices associated with $\alpha$ T-GCs of $\rho^{\iota}$ and $\Tilde{P}^{\iota} = Blkdig(P^{\iota}_1, P^{\iota}_2, \cdots, P^{\iota}_{\alpha})$. 

\subsubsection{\textbf{RCGMC}}
\label{subsubsec: agent centric GM}
The following MISDP models our robot-centric geometric mean configuration generation approach: 
\vspace{-0.3in}
\begin{align}
	\label{eqn: agent GM MISDP obj}
	\minimize_{\substack{{A} \in \mathcal{S}^n_+,\ \nu \in \R_{> 0}, \\ {\Pi} \in \{0,1\}^{n\times n}}} \quad &  - {{0}}_{\Bar{\iota}}^{1\times n} {A}  \begin{bmatrix}
\frac{1}{\alpha}\text{Trace}((\Tilde{P}^{1})^{-1})\\ 
\frac{1}{\alpha}\text{Trace}((\Tilde{P}^{2})^{-1})\\
\vdots\\
\frac{1}{\alpha}\text{Trace}((\Tilde{P}^{n})^{-1})
\end{bmatrix}\\
	\text{subject to} ~~ & \label{eqn:cons:unity sum}
	{A}\cdot\bar{1}^n = \bar{1}^n\\
	\label{eqn:cons:connectivity}
	~~ &\frac{1}{n} \Bar{1} \Bar{1}^T  + (1 - \nu) {I}  \succeq {A}, \nu  \ll  1  \\
	\label{eqn:cons:diag binary}
	~~ & diag({\Pi}) = \Bar{1}^n\ \\
	\label{eqn:cons:binary sym}
	~~ & {\Pi} = {\Pi}^T\\
	\label{eqn:cons:adj diag}
	~~ & [{A}]_{i,i} > 0 ~ \forall\ i \in [n]\\
	\label{eqn:cons:adj off diag min}
	~~ & [{A}]_{i,j} \geq 0  \forall~(i, j) \in [n]^2,~i \neq j \\
	\label{eqn:cons:adj off diag max}
	~~ & [{A}]_{i,j} \leq  {\Pi}_{i,j} \forall ~(i, j) \in [n]^2,~i \neq j\\
	\label{eqn:cons:topology near}
	~~ & \|{\Pi} - {{A}_u}[f_p] \|_F^2 \leq 2 \times e.
\end{align}

The decision variables ${A}$ and ${\Pi}$ model the doubly stochastic matrix used for the consensus protocol and the adjacency matrix of the generate configuration respectively. \hyperref[eqn:cons:unity sum]{Constraint~\ref{eqn:cons:unity sum}} and  \hyperref[eqn:cons:adj diag]{Constraint~\ref{eqn:cons:adj diag}}
 to \hyperref[eqn:cons:adj off diag max]{Constraint~\ref{eqn:cons:adj off diag max}} ensures that ${A}$ is a doubly stochastic matrix that is structurally equivalent to ${\Pi}$. In the light of \autoref{theorem:graph connectivity}, \hyperref[eqn:cons:connectivity]{Constraint~\ref{eqn:cons:connectivity}} enforces the generated configuration to possess a connected graph. Finally, \hyperref[eqn:cons:topology near]{Constraint~\ref{eqn:cons:topology near}} encodes the near topology condition (condition 2) in \autoref{prob: configuration generation} into the MISDP. If $\iota$ represents the label of the robot whose sensor quality deteriorated at $f_p$, then with some simple algebraic manipulation it can be easily shown that \autoref{eqn: agent GM MISDP obj} results in the average over the trace of the fused GCs according to \autoref{eqn: GMF covariance rule}.

\subsubsection{\textbf{TCGMC}}
\label{subsubsec: team centric GM}

Consider the following MISDP formulation encoding the team-centric geometric mean configuration generation strategy.
%
\begin{align}
	\label{eqn: team GM MISDP obj}
	\minimize_{\substack{{A} \in \mathcal{S}^n_+,\ \nu \in \R_{> 0}, \\ {\Pi} \in \{0,1\}^{n\times n} \\
	{\Bar{P}, \Bar{\Delta}} \in \mathcal{S}^{n\times s_a \times \alpha}_+}}\  \quad & \text{Trace}({\Bar{P}}) \\
	\text{subject to} ~~ 
	\label{eqn:cons:schur complement}
	\begin{bmatrix}
	{\Bar{P}} & {I} \\
	{I} &  {\Bar{\Delta}}
	\end{bmatrix} &~ \succeq 0 \\
	\label{eqn:cons: GM knone}
	~~ {A} \otimes {I}\begin{bmatrix}
{(\Tilde{P}^{1})^{-1}}\\ 
{(\Tilde{P}^{2})^{-1}}\\
\vdots\\
{(\Tilde{P}^{n})^{-1}} 
\end{bmatrix} &= \begin{bmatrix} 
{\Delta_1}\\ 
{\Delta_2}\\
\vdots\\
{\Delta_n} 
\end{bmatrix}\\ 
~~ \text{\hyperref[eqn:cons:unity sum]{Constraint~\ref{eqn:cons:unity sum}}} &-  \text{\hyperref[eqn:cons:topology near]{Constraint~\ref{eqn:cons:topology near}}}.\nonumber
\end{align}



Where ${\Bar{\Delta}} = Blkdig({\Delta}_1, {\Delta}_2, \cdots, {\Delta}_n)$ and ${A} \otimes {I}$ results in the \textit{Kronecker product} \cite{Horn:1985:MA:5509} between ${A}$ and the identity matrix which matches the size of $\Tilde{P}^{\iota}$. \hyperref[eqn:cons: GM knone]{Constraint~\ref{eqn:cons: GM knone}} is the covariance fusion rule \autoref{eqn: GMF covariance rule} for the whole team written compactly as a single equation. In addition,  
\hyperref[eqn:cons:unity sum]{Constraints~\ref{eqn:cons:unity sum}}-\ref{eqn:cons:topology near} are also required for \textbf{TCGMC}.  The following lemma proves that minimizing \autoref{eqn: team GM MISDP obj}  minimizes $ \frac{1}{n\times \alpha} \sum_i^{n} \text{Trace}({P}^{i,\alpha}_{gm}))$, where ${P}^{i,\alpha}_{gm}$ is the block diagonal matrix containing $\alpha$ number of GMF fused GCs' covariance matrices associated with the $i^{th}$ tracker (See  \hyperref[app: proof lemma]{Appendix~\ref{app: proof lemma}} for the proof). 
\begin{lemma}
\label{lemma: trace bound}
The  $\frac{1}{n\times \alpha} Tr(\Bar{{P}})$ is an upper bound on $\frac{1}{n \times \alpha} \sum_i^{n} \text{Trace}({P}^{i,\alpha}_{gm})$
\end{lemma}
%
%
\subsubsection{\textbf{RCAMC}}
\label{subsubsec: agent centric AM}

Similar to \autoref{subsubsec: agent centric GM}, we formulated the MISDP for robot-centric arithmetic mean configuration generation as:
\vspace{-0.25in}
\begin{align}
\label{eqn: agent AM MISDP obj}
    \minimize_{\substack{{A} \in \mathcal{S}^n_+,\ \nu \in \R_{> 0}, \\ {\Pi} \in \{0,1\}^{n\times n}}} \quad &   {{0}}_{\Bar{\iota}}^{1\times n} {A}  \begin{bmatrix}
\frac{1}{Ws}\text{Trace}((\Tilde{P}^{1}))\\ 
\frac{1}{Ws}\text{Trace}((\Tilde{P}^{2}))\\
\vdots\\
\frac{1}{Ws}\text{Trace}((\Tilde{P}^{n}))
\end{bmatrix} \\
\text{subject to} ~~ 
~~ \text{\hyperref[eqn:cons:unity sum]{Constraint~\ref{eqn:cons:unity sum}}} &-  \text{\hyperref[eqn:cons:topology near]{Constraint~\ref{eqn:cons:topology near}}}.\nonumber
\end{align}

Here the objective function \autoref{eqn: agent AM MISDP obj} is a direct result of the application of \autoref{eqn: AMF covariance rule} for $\rho^{\iota}$. 

\subsubsection{\textbf{TCAMC}}
\label{subsubsec: team centric AM}

Finally, the MISDP formulation for the team-centric arithmetic mean configuration generation can be expressed as: 
\vspace{-0.2in}
\begin{align}
    \minimize_{\substack{{A} \in \mathcal{S}^n_+,\ \nu \in \R_{> 0}, \\ {\Pi} \in \{0,1\}^{n\times n} \\
	{\Bar{P}, \Bar{\Delta}} \in \mathcal{S}^{n\times s_a \times \alpha}_+}}\  \quad & \text{Trace}\left(\Bar{\Delta}\right) \\
	\text{subject to} 
~~& {A} \otimes {I}\begin{bmatrix}
(\Tilde{P}^{1})\\ 
(\Tilde{P}^{2})\\
\vdots\\
(\Tilde{P}^{n}) 
\end{bmatrix} = \begin{bmatrix} 
{\Delta_1}\\ 
{\Delta_2}\\
\vdots\\
{\Delta_n} 
\end{bmatrix}\\
~~ \text{\hyperref[eqn:cons:unity sum]{Constraint~\ref{eqn:cons:unity sum}}} &-  \text{\hyperref[eqn:cons:topology near]{Constraint~\ref{eqn:cons:topology near}}}.\nonumber
\end{align}

Where $\Bar{\Delta} = Blkdig(\Delta_1, \Delta_2, \cdots, \Delta_n)$ and therefore minimizing the trace of $\Bar{\Delta}$ results in minimizing sum of the traces of AMF fused covariance matrices of the trackers in the team. 

\subsection{Formation Synthesis}
\label{subsec: formation syn}

Once a new configuration is generated, we assign a physical location to each robot to maximize the team's non-overlapping coverage of the space.  In this assignment problem, we impose constraints to ensure that connected robot pairs remain within communication distance $d_{mc}$ of each other, and that the distance between all robot pairs exceeds $d_s$ to avoid collision.  An additional constraint is added to ensure that each robot is no more than $E$ distance away from the centroid of the {T-GC}s.




This produces the following constrained optimization problem:
%
\begin{align}
    \label{obj-coverage}
    \hspace{-0.1in}
	\max_{\{X_{|n|}\}} \pi \sum_{i \in \cV} \left(({d^i_{sen}})^2 \hspace{-0.05in} - \hspace{-0.05in} \sum_{j \in \cV \neq i}\frac{(2d^i_{sen} - \|X_i - X_j\|)^2}{2} \right)
\end{align}
\begin{align}
	\label{con-collision}
	\vspace{-0.3in}
	\text{subject to} \quad & d_s \leq \|X_i - X_j\| \ \leq d_{mc} & & \forall\ (i, j) \in \cE \\
	\label{con-nonneighbor}
	& d_{s} \leq \|X_i - X_j\| & & \forall\ (i, j) \in \overline{\cE} \\
	\label{con-box}
	& B^{\min} \leq X_i \leq B^{\max} & & \forall\ i \in V \\
	\label{teamdist2target}
	& \|X_{team} - X_{target}\| \ \leq E & &
	\end{align}
	where $d^i_{sen}$ is the radius of the circular field of vision of tracker, $\rho^i$,
	$X_{team}$ is the average position of the robot team, $X_{target}$ is the centroid of the {T-GC}s of the failure node, $E$ is the user-defined maximum distance the team can be from $X_{target}$,
	and $B^{\min},\ B^{\max} \in \R^3$ are the minimum and maximum extents of an axis-aligned bounding box, with the operator $\leq$ applied elementwise in~\autoref{con-box}. 
We solve the formation synthesis optimization problem \autoref{obj-coverage} - \autoref{teamdist2target} following the simulated annealing approach described in \cite{ramachandran2019resilience}.


\begin{figure}[t]
	\centering
	\begin{tabular}{c}
		\subcaptionbox{Before Failure (Overhead)}{\includegraphics[width=.75\linewidth]{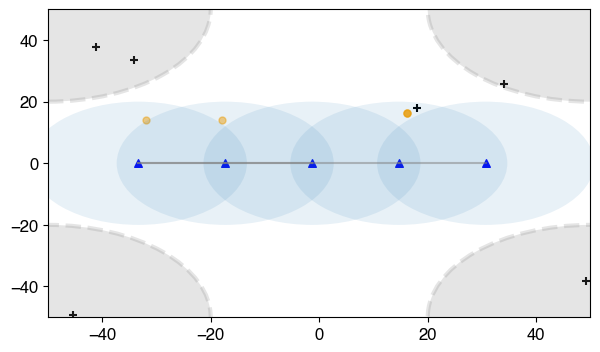}}
		 \\ 
		\subcaptionbox{After Failure (Overhead)}{\includegraphics[width=.75\linewidth]{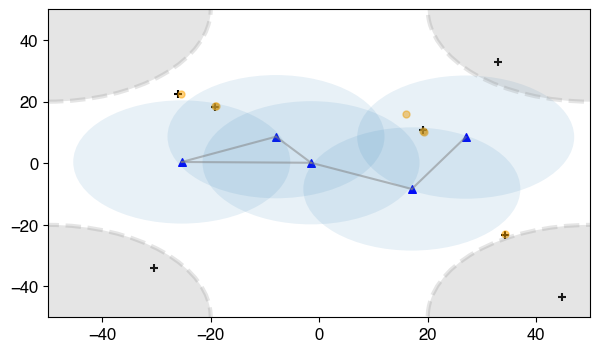}}
	\end{tabular}
	\caption{\small{Screenshots of a simulation in which a robot team of five tracks targets moving below them (overhead view). A robot's sensing area is depicted as a light blue circle.  The target birth areas are depicted as light gray circles in the corners. The true target positions are denoted as black '+'s.  The target position estimates are denoted as orange circles.  The robots themselves are denoted as dark blue triangles.  The figure on the left depicts the formation before the occurrence of a sensor deterioration event. The corresponding figures on the right portrays the formation after 1) sensor deterioration is detected, 2) a new communication edge is chosen, and 3) the robots move to their new locations.}}  
	\label{fig:sim failures}
	\vspace{-0.25in}
\end{figure}

\begin{figure}[t]
	\centering
	\begin{tabular}{c}
		{\includegraphics[width=.8\linewidth]{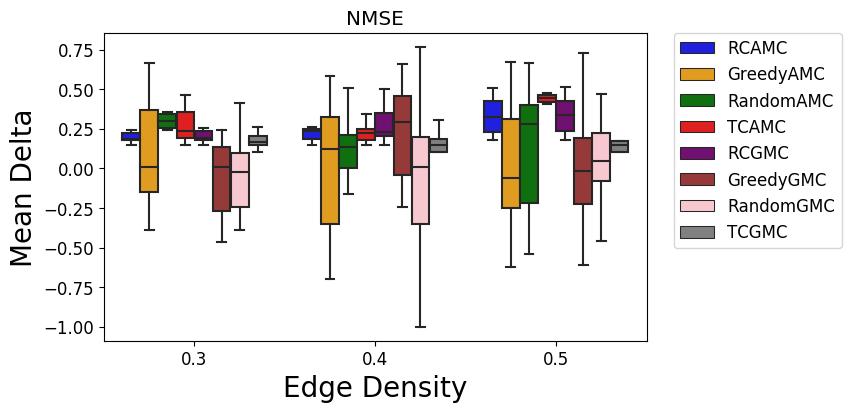}}
		\\
	    {\includegraphics[width=.8\linewidth]{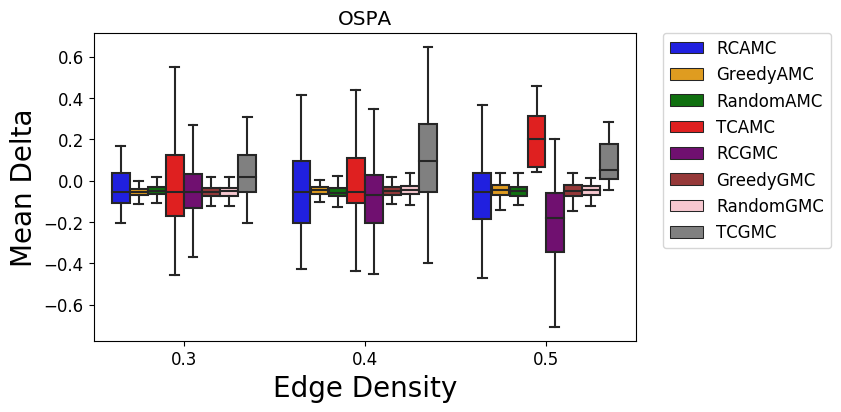}}
	\end{tabular}
	\caption{\small{The average difference (normalized between -1 and 1) in NMSE and OSPA between the \textit{baseline} scenario and each configuration strategy at different graph edge densities.  Note that all robot team sizes are aggregated here, causing larger variance of OSPA for the team-centric and agent-centric strategies than the greedy and random strategies.  In contrast, the greedy and random strategies have larger variance in NMSE than the team-centric and agent-centric strategies.  Across all edge density levels simulated, the team-centric strategies consistently perform best in maximizing the difference for both metrics.
	}}
	\label{fig:sim validation nmse ospa}
	\vspace{-0.25in}
\end{figure}


\section{Simulation}
\label{sec:simulation}

To validate our approach, we conducted multiple simulations of a robot team tracking multiple moving targets.  A target would be born in a random corner of 2D space with bounding box of $x \in [-50, 50]$, $y \in [-50, 50]$.  The target would then move in a straight line trajectory to the opposite corner of the bounding box.  The number of new targets born at each time step was determined by a random Poisson point process with $\lambda = 1$.  
The initial position for a new born target was chosen according to a Gaussian distribution about a 20 unit radius of each corner.  
\autoref{fig:sim failures} illustrates this target birth area.

For the local GM-PHD filter, we set the survival probability of each target to 0.98, and detection probability of each target to 0.95 if within the field of vision of a robot, and 0 otherwise.  Targets within Mahalanobis distance of 0.2 were merged together at each local GM-PHD filter iteration and during fusion.  Targets with weights less than 1e-6 were pruned.

We initiated the same ${H}^{\iota}_k$ and ${R}^{\iota}_k$ for each robot in the tracker team.  We used $L=n/2$ for the consensus step, where $n$ is the size of the team.  Parameters chosen for the configuration generation and formation synthesis problems were 
$n_e = 1$, $d_{s}=10$, $d_{mc}=25$, and $d^{\iota}_{sen} = 20  \forall\ {\iota} \in [n]$, with a bounding box of $x \in [-50, 50]$, $y \in [-50, 100]$, and $z \in [0, 100]$.

To simulate deteriorating sensor quality for $\rho^{\iota}$, we modified its covariance matrix ${R}^{\iota}_k$ by adding a random positive definite matrix.  We generated various deterioration event sequences for robot teams of $n \in \{5, 6, 7, 10, 12, 15, 20, 25, 30\}$  where a random robot was chosen at every \textit{f} time step of the simulation to experience sensor deterioration.  
\autoref{fig:sim failures} shows an overhead view of a single simulation trial with 5 robots.  

For all configuration generation approaches, we simulated 30 deterioration sequences for each size robot team.  
We compare each configuration generation approach with a \textit{baseline} scenario in which no edges are added at failure.  Each trial was initialized with a line graph.
The target dynamics and distributed GM-PHD filter were implemented in Python. For the agent-centric and team-centric approaches, the MISDP problem was solved using Python with PICOS as the optimization problem modeling interface and MOSEK as the semi-definite programming solver.  
In both approaches, the simulated annealing technique for formation synthesis was implemented in Python.  

All simulation computations were performed on a 64-bit Ubuntu 18.04 desktop with 3GHz  Intel Core Xeon Gold 6154 CPU and 256 GB RAM. Additionally, we employed GNU-Parallel 
to parallelize our computations on this machine.

To quantify performance, we used the the optimal subpattern assignment (OSPA) distance \cite{schuhmacher-ospa} to evaluate the estimation error of the target positions after PHD fusion.  The OSPA metric represents the distance between two sets.  In our case, this is the distance between the set of the true target positions and the set of the {T-GC}s means.  For the OSPA calculation, we use cutoff parameter $c$ = 5 and order parameter $p$ = 1.

Additionally, we evaluate the target set cardinality estimate of the team using the normalized mean squared error (NMSE). The results are presented in \autoref{fig:sim validation nmse ospa}. 
%

Over all our trials, the team-centric and robot-centric strategies did not exceed 60\% edge density, while this was quickly exceeded in the greedy and random strategies.  This means that often readjusting network weights is enough to improve sensing quality of the network and increasing connectivity between robots is not necessary.

While we did not directly optimize for NMSE or OSPA in our approach, both proposed team-centric and robot-centric strategies, using either arithmetic mean or geometric mean fusion, on average perform better in minimizing these metrics than the random and greedy strategies.  Between the team-centric and robot-centric strategies, the team-centric strategies perform best.

\section{Conclusion}
\label{sec:conclusion}

This paper presents a novel strategy that facilitates a team of robots performing multi-target tracking to respond to a sensor fault in one of the team members by reconfiguring the team's communication network. The reconfigured team attenuates the adverse effect of sensor quality deterioration on multi-target tracking performance of the team. We presented four different MISDP formulations to compute the new robot team configuration. All formulation were validated in simulation and compared to each other.  In future, we plan to validate our approach on our multi-robot testbed~\cite{crazyswarm}.


\appendices
\section{}
\label{app: graph connect proof}
\textbf{Proof of \autoref{theorem:graph connectivity}:} Let ${L} = {I}-{A}$, then since ${A}$ is doubly stochastic ${L}\bar{1}^n = \bar{0}^n$ and ${L}^T\bar{1}^n = \bar{0}^n$. Also, as the spectrum of ${A}$ is real and less than or equal to one in magnitude, the spectrum of ${L}$ is real and less than or equal to zero. Now, from the above statement we conclude that ${L}$ is a positive semi-definite matrix. Furthermore, note that ${L}$ can be interpreted as the Laplacian of a weighted undirected graph $\mathcal{G}_L$ having the same topology of the graph associated with ${A}$ except for self loops. Since the connectivity properties of an undirected graph does not depend on the existence of self loops, original graph(graph associated with ${A}$) is connected if and only if   $\mathcal{G}_L$ is connected. From ~\cite[Proposition 1]{sundin2017connectedness}, we infer that $\mathcal{G}_L$ is connected if and only if ${L} + \frac{1}{n}\bar{1}\bar{1}^{\top} \succ 0$.  Therefore, substituting ${L} = {I}-{A}$ in the equation yields $\frac{1}{n} \bar{1} \bar{1}^{\top}  + {I}  \succ {A}$.

\section{}
\label{app: proof lemma}
\textbf{Proof of \autoref{lemma: trace bound}:} According to Schur complement lemma \cite[Chapter 2]{boyd1994linear}, the following \textit{linear matrix inequality}(LMI)\cite{boyd1994linear}, $\begin{bmatrix}
    {Q} & {S} \\
    {S}^T & {R}
    \end{bmatrix} \succeq 0$
%
%
is equivalent to ${R} \succeq 0$, ${Q} - {S}{R}^{-1}{S}^T \succeq 0$. Therefore, the LMI \autoref{eqn:cons:schur complement} is equivalent to ${\Bar{P}} - {\Bar{\Delta}}^{-1} \succeq 0 $. Also, it is straightforward to see that Trace(${\Bar{P}}$) $\geq$ $Tr({\Bar{\Delta}}^{-1})$. Since, ${\Bar{\Delta}}$ is the block diagonal matrix containing the posterior information matrices of all tracking robots, ${\Bar{\Delta}}^{-1}$ is a block diagonal matrix with $\{{P}^{1,\alpha}_{gm}, {P}^{2,\alpha}_{gm}, \cdots, {P}^{n,\alpha}_{gm}\}$ along its diagonal.
Therefore,  $\frac{1}{n\times \alpha}Tr({\Bar{P}})$ $\geq$ $\frac{1}{n\times \alpha}Tr({\Bar{\Delta}}^{-1})$ is equivalent to $\frac{1}{n \times \alpha}Tr({\Bar{P}})$ $\geq$  $\frac{1}{n}\sum_i^{n \times \alpha} Tr({P}^{i,\alpha}_{gm})$


\end{document}